# DISK FAILURE PREDICTION BASED ON MULTI-LAYER DOMAIN ADAPTIVE LEARNING

Guangfu Gao [1], Peng Wu[1] and Hussain Dawood [2]

[1] School of lnformation Science and Engineering, University of Jinan, Jinan, China
ggf66885213@163.com, ise_wup@ujn.edu.cn

[2] Department of Information Engineering Technology, National Skills University Islamabad, Islamabad
Hussain.dawood@nsu.edu.pk

## ABSTRACT

*Large scale data storage is susceptible to failure. As disks are damaged and replaced, traditional machine learning models, which rely on historical data to make predictions, struggle to accurately predict disk failures. This paper presents a novel method for predicting disk failures by leveraging multi-layer domain adaptive learning techniques. First, disk data with numerous faults is selected as the source domain, and disk data with fewer faults is selected as the target domain. A training of the feature extraction network is performed with the selected origin and destination domains. The contrast between the two domains facilitates the transfer of diagnostic knowledge from the domain of source and target. According to the experimental findings, it has been demonstrated that the proposed technique can generate a reliable prediction model and improve the ability to predict failures on disk data with few failure samples.*

## 1. INTRODUCTION

In today's era of big data, disk is the most common and widely used storage device. Despite its longevity, the disk is vulnerable to external environmental factors such as temperature, humidity, and collisions, and disk failures have become the norm. The study of predicting disk failures becomes essential due to the imperative need to avoid significant economic losses resulting from data loss. Upon the event of a fault, data recovery becomes unfeasible without a prior backup procedure, which entails intricate consequences and ramifications. Therefore, it is crucial to delve into the realm of disk fault prediction.

The progression from passive fault tolerance methods, aimed at enhancing system reliability through duplication or code correction techniques, towards active fault tolerance SMART techniques for disks involves the monitoring of attribute value fluctuations. When the threshold value is surpassed, it indicates an impending failure within a specific timeframe. While data backups ensure security and reliability, their effectiveness is hindered by low accuracy rates. Researchers have conducted many studies based on the measured values of each attribute. Wang Yu et al. [1] presented a disk detection approach utilizing the Mahalanobis Distance (MD). Zhu Bing-peng et al. [2] employed both back-propagation neural network and an improved support vector machine model for disk failure prediction. To address experimental false positives, Jia Runying et al. [3] utilized the Adaboost algorithm to amalgamate BP neural networks. LSTM neural networks were utilized by Lima et al. [4] and Kang Yanlong [5] to achieve relatively favorable results in disk failure prediction. To tackle imbalanced disk data, Li Xinpeng et al. [6] proposed an enhanced Bagging-GB-DT algorithm for future data prediction. Liu Xin et al. [7] analyzed critical factors impacting disk failure using the Shapley Additive exPlanations (SHAP) model, providing advice for subsequent disk prediction. Zhu Hongbin et al. [8] developed a failure prognostication technique well-suited for household storage appliances, utilizing deep learning algorithms and validating its efficacy. Deng Ling et al. [9] examined the SMART data and used the random forest predictor to predict the remaining life of the disc. Based on the Light Gradient Boosting Machine (LightGBM) method, Tian Xuanyu et al. [10] proposed a model that can predict disk failures online, achieved a high failure detection rate, and solved the aging problem of the model by updating data online. To address the issue of data imbalance, one can implement solutions at the data level, thereby compensating for data insufficiency and enabling machine learning algorithms to attain accurate predictions. Jia et al. [11] employed the Conditional Tabular Generative Adversarial Networks (CTGAN) approach to tackle the issue of imbalance in disk data by generating new synthetic data at the data level. By combining the different GANs'advantages to generate data, Yuan et al. [12] used the ensemble learning method, then applied machine learning and neural network techniques to the prediction, ultimately producing a superior model.

Traditional machine learning algorithms are trained on a large amount of disk failure data. As disks are damaged and replaced, the distribution of data between different types of disks is different, and due to the small amount of data, the model obtained from traditional machine learning training often fails to make good predictions in similar new areas. In many fields where machine learning is applied, the assumption of an independent and identical distribution of data often does not hold. Transfer learning and domain adaptation techniques can improve the effectiveness of machine learning models in cross-domain tasks. To eliminate the train-test mismatch for heterogeneous disks, we use domain-adaptive learning in transfer learning to build a well-established model for predicting disk failure. Using local maximum difference and other measures, we align the origin and destination domains in the common characteristic space, thereby improving the generalisation performance of the model, as mentioned in the paper [13][14][15].

References [16][17] describe many representative methods based on data and models in transfer learning and domain adaptation. Reference [18] describes a review of domain adaptation methods, based on domain distribution differences, confrontation, reconstruction. According to the literatures [19][20], we understand the relevant software reliability modeling foundation, standardize the current experiment, and obtain a model with high predictive performance, which is convenient for subsequent description with an approximate mathematical model. Domain adaptation and transfer learning have achieved significant maturity in various computer vision domains, encompassing picture categorization, goal identification, video analysis, and more. These techniques have also found extensive applications in domains such as medical imaging, transportation, recommendation systems, and others. However, their utilization in the field of disk failure prediction remains limited.

In this paper, we introduce a transfer learning approach called Multi-layer Domain Adaptation (MDA) method. It leverages abundant previous data from intact models and limited fault samples from models with fewer failures for training purposes. At each layer, we incorporate maximum mean difference and other relevant adaptation metrics to enhance the performance of the proposed method. A transfer learning model is trained and constructed, efforts are undertaken to diminish the discrepancy in data distribution between the source and target domains, and the failure prediction of disc data with few faulty samples is improved.

The following are key contributions of this research:

1. Based on the domain-adaptive framework, make full use of deeper features instead of single-layer features.

2. Based on the classification loss, adding the maximum mean difference and correlation alignment together reduce the distribution difference between domains.

The remaining article is structured as follows: 'Section 2' briefly introduces the theoretical background of transfer learning and domain adaptation. "Section 3" provides the experimental procedure. "Section 4" provides the related work and details of the methods. "Section 5 and discussions" provides an analysis and its comparison with other approaches. "Section 6" concludes the research with some future recommendations.

This document describes, and is written to conform to, author guidelines for the journals of AIRCC series. It is prepared in Microsoft Word as a .doc document. Although other means of preparation are acceptable, final, camera-ready versions must conform to this layout. Microsoft Word terminology is used where appropriate in this document. Although formatting instructions may often appear daunting, the simplest approach is to use this template and insert headings and text into it as appropriate.

## 2. RELATED WORK

### 2.1. Transfer learning

Transfer learning, a technique in machine learning, utilises existing knowledge in other domains where relevant information is scarce, thus alleviating the problem of insufficient data in the target domain. The objective of transfer learning is to generalize knowledge acquired in a specific domain to effectively address problems in other related or similar domains. This technique capitalizes on the shared characteristics between the domains to enable efficient knowledge transfer. Transfer learning has many application scenarios. It has already been used in literature [21][22][23][24][25][26] to detect crop pests, medical diseases, defects in bamboo leaves and faults in electrical loads. The predictive performance of the model can be greatly enhanced by transfer learning, which is far more effective than traditional machine learning methods when the current samples are insufficient. In the realm of disk failure prevention, there has been a growing adoption of transfer learning techniques in recent times. Zhang et al. [27] proposed a transfer learning method of TLDFP for disk failure prediction, using an iterative algorithm to update the weights and give more weight to hard to classify instances. Firstly, the Kullback-Leibler Divergence (KLD) value is used to select the source domain, and Jensen-Shannon Divergence (JSD) and Wasserstein Distance (WD) are also tested, which proves that the KLD method has better effectiveness. Gao et al. [28] first proposed using the maximum mean difference as an indicator for selecting the appropriate majority of disc model data as the source domain, and verified the effectiveness of their method.

### 2.2. Domain adaptation

Given the scarcity of labeled data in the target domain, where a substantial volume is not available and a high performance machine learning model cannot be taught, model pre-training can be considered in an analogous but different auxiliary domain where a significant amount of labelled data can be obtained. Subsequently, by fine-tuning the pre-existing model to suit the unique characteristics of target domain, the practical challenge of insufficient fault data in target domain can be effectively addressed. However, the different distribution of data across domains becomes an obstacle to model transfer. The acquisition of a general model is aimed at adapting to a domain, so that the knowledge gained in a domain with abundant data can be applied more efficiently, thereby mitigating the challenge posed by limited data in the target domain. By using domain adaptation strategies, the disparity in data distribution between the two domains can be minimised, allowing domain-invariant knowledge to be transferred and reused between them. Zhao et al. [29] initially introduced domain adaptation to train a predictive model for disk failure, employing a common feature representation across different domains. Jiang et al. [30] transform 1D SMART attributes into 2D attributes as input for GAN's disk failure prediction method to process data, adopt 2D convolutional neural network as feature lifter, and combine classifier and domain discriminator to train classification, predict the probability of failure over a period of time.

Our method incorporates Maximum Mean Difference (MMD) and Correlation Alignment (CORAL) in multi-layer domain adaptation as a metric loss to effectively minimize the distribution gap between the source and target domains, which can achieve better prediction results.

## 3. EXPERIMENTS

### 3.1. Experimental data

The dataset used in this study is sourced from the disk data published by Blackblaze in 2021. Blackblaze collected statistics from their data center, encompassing 170,000 disks and 255 SMART attributes. For the purposes of this paper, we have specifically chosen 9 attributes that are relevant to disk failure prediction. These attributes include the initial values of two specific attributes and various types of disks that had the highest failure rates. This is a number of failed disks, such as ST4000DM000, ST8000DM002, ST12000NM001G, ST12000NM0007, ST8000NM0055, ST12000NM0008, ST500LM030, ST14000NM001G, ST14000NM0138, ST18000NM000J and ST10000NM0086. Based on the datasets consisting of affirmative and adverse samples, the construction of the source and target domains is carried out using a ratio of 1:10.

The 11 data items in table 1 are most closely related to changes in disk status among the selected SMART data after data analysis. The ID in the table is the original ID number of the data item in SMART, and the specific name of each selected data item is given. In particular, it should be noted that in SMART, each item is represented by two numerical forms, namely normal value and raw value. In the training data, the normal value is used for almost all data elements, except for 5 and 197. As the data analysis shows that their raw value may actually be more sensitive to changes in disc status, both raw and normal values are used for these two items.

Table 1 Part ID codes and attribute names in SMART.

| Attributes ID | Attributes Name | Attributes Meaning |
|---|---|---|
| 1 | read error rate | Low-level data read error rate |
| 3 | Spin-Up Time | Disc boot time |
| 5 | Real located Sector Count | Relocation Sector Count |
| 5 (RAW) | Real located Sector Count | Relocation Sector Count |
| 7 | Seek Error Rate | Seek error rate |
| 9 | Power-On Hours | Disk power up time |
| 187 | Reported Uncorrectable Errors | Report uncorrectable errors |
| 189 | High Fly Writes | Head write height |
| 194 | Temperature | Temperature |
| 197 | Current Pending Sector Count | Pending reset sector count |
| 197 (RAW) | Current Pending Sector Count | Pending reset sector count |

### 3.2. Data preprocessing

To prepare the extracted dataset for further analysis, it is crucial to perform eigenvalue normalization. This process ensures that the data is dimensionless and confined within the range of [-1,1]. To achieve this, a straightforward linear transformation function is employed for data normalization. Equation (1) outlines the specific definition and calculation of the normalization process.

$$x_{normal} = 2 * \frac{x - x_{min}}{x_{max} - x_{min}} - 1 \tag{1}$$

The original value of the data is denoted by $x$ and $x_{normal}$ is normalized by the equation. The equation's $x_{max}$ and $x_{min}$ are the maximum and minimum values of its current features, respectively.

### 3.3. Evaluation indicators

To evaluate the model's reliability and accuracy in predicting disc failure, the disc failure samples are considered positive samples (labelled P) and the good disc samples are considered negative samples (labelled N). True (labelled T), False (labelled F). TP in Table 2 indicates that the model predicts the failed disk correctly. FP means the model is incorrectly predicting good samples. FN means the model predicted the failed disc incorrectly. TN means that the model correctly predicts good samples. G-mean is used based on the confusion matrix as shown in equation (2). When dealing with unbalanced affirmative and opposing samples, it makes more sense than accuracy and recall metrics.

Table 2 Confusion Matrix.

| Category | Predicted failure | Predicted anormal |
|---|---|---|
| Actually failure | TP | FN |
| Actually normal | FP | TN |

$$G - mean = \sqrt{\frac{TP}{TP + FN} * \frac{TN}{TN + FP}} \tag{2}$$

## 4. METHODS

### 4.1. Network structure

A fault prediction module and a domain adaptation module form the multi-layer domain adaptation network. The domain adaptation module mainly reduces the difference in data distribution between the two domains, allowing the feature extractor to learn general invariant features. The failure prediction module mainly uses the learned invariant features to predict disk failures. The domain adaptation module consists mainly of a multi-layer metric difference structure. Figure 1 illustrates the network architecture incorporating maximum distribution disparity and correlation alignment for aligning the two domains, consequently reducing the distribution difference between them.

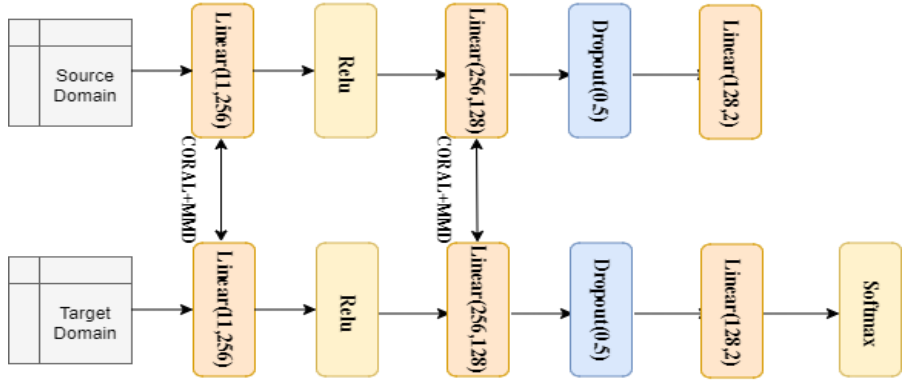

Figure 1. MDA network structure

**4.1.1. Failure prediction module**

The fault prediction module is composed of five elements: a data input segment, two fully connected segments, two activation segments, a dropout segment, and a softmax output classification segment. Of these, the output layer uses the softmax. Figure 1 presents the network's main parameters. When checking the neuron count in the fully connected layer, experiments have yielded 64, 128, 256, 648 and 1024, with 256 being the highest. For the two-layer activation layers, we chose the ReLu activation function. According to the literatures [31][32] Sigmoid function approximates the step function, which is also suitable for binary classification. In our experiment, we employ the Rectified Linear Unit (ReLU) activation function, which effectively addresses the issue of gradient vanishing. One-sided saturation can also make neurons more robust to noise interference. Using the ReLU activation function is also computationally efficient. And ReLU truncates negative values to 0, which introduces sparsity into the network and further improves computational efficiency. Then Dropout is used to avoid the problem of overfitting according to the structure of first big and then small, and the loss rate of neurons is set to 0.5.

**4.1.2. Domain Adaptation module**

To mitigate the divergence between distributions, the domain adaptation module aims to minimize the dissimilarity by using the maximum mean difference and the corresponding orientation to measure the second order statistic (covariance) of the multi-layer features. This is accomplished by comparing MMD with only a single layer, CORAL with a single layer, and MMD with a double layer, CORAL for two layers, and MMD and CORAL for single layers to minimize distribution differences.

Domain adaptation often relies on the utilization of loss functions, with the MMD being a widely favored choice in transfer learning scenarios. The equation (3) defines the distance between two domain distributions that are different but related.

$$\text{MMD} = || \frac{1}{n}\sum_{i=1}^{n} Q(X_i) - \frac{1}{m}\sum_{j=1}^{m} Q(Y_i) ||_H^2 \qquad (3)$$

where $Q()$ maps the data to the Reproducing Kernel Hilbert Space (RKHS), denoted by H.

Let the training samples in the source domain $D_S=\{x_i\}$, $x \in \mathbb{R}^d$, and their labels are $L_S = y_i$, $i \in$ (1, ..., L), the unlabeled target domain data $D_T = \{u_i\}$, $u \in \mathbb{R}^d$, where $d$ can be understood as the output number of the full connection layer of the network, that is, the number of neurons in the full connection layer. $D_S^{ij} D_T^{ij}$ represents the vector samples of the $j$th dimension of the $i$th source domain (target domain), $C_S C_T$ represents the feature covariance matrix.

Equation 4 illustrates the CORAL loss as the gap between two second-order domain statistics.

$$l_{CORAL} = \frac{1}{4d^2}||C_S - C_T||_F^2 \tag{4}$$

where $||\cdot||_F^2$ represents the Frobenius norm. From this, the data covariance of source and target is obtained as shown in equations (5) and (6).

$$C_S = \frac{1}{n_S - 1}\left(D_S^T D_S - \frac{1}{n_S}(\mathbf{1}^T D_S)^T(\mathbf{1}^T D_S)\right) \tag{5}$$

$$C_T = \frac{1}{n_T - 1}\left(D_T^T D_T - \frac{1}{n_T}(\mathbf{1}^T D_T)^T(\mathbf{1}^T D_T)\right) \tag{6}$$

For the proposed multi-layer domain adaptation, MMD and CORAL are used to measure the loss of each layer and update the weights in real time. Equations (5) and (6), the feature covariance matrices of the source domain $D_S$ and the target domain $D_T$, are given by a column vector $\mathbf{1}$ with all elements equal to 1. The source and target domains' layer features are denoted by $C_S$ and $C_T$, respectively.

### 4.2. Optimization target

The MDA network has the following two optimization objectives:

1) Minimize loss of classification for failure classes on the origin domain dataset. To ensure accurate identification of the disk's health status and proper training of the MDA network, calculating the classification loss is crucial. This loss can be expressed as the standard softmax classification loss in cross-entropy loss.

2) In order to minimize the discrepancies between the source and target domains, various methods for domain adaptation have been proposed. These methods aim to reduce the MMD and CORAL metric losses by considering multi-layer features from both domains. In addition, common features are extracted. To ensure model accuracy, the parameters of the network model can be updated while jointly minimizing the two-layer densely connected network by combining the aforementioned losses: classification loss, MMD loss, and CORAL loss. Generalization performance on, resulting in a stable model whose total loss is defined as equation (7).

$$l_{LOSS} = n * l_{CLASS} + \sum_{i=1}^{t} x_i l_{MMD}^i + \sum_{i=1}^{t} y_i l_{CORAL}^i \tag{7}$$

where $t$ represents the number of layers in the network that need to adapt to the CORAL loss; $x_i$ and $y_i$ are the domain adaptation loss weights under the corresponding layers; $n$ is the weight corresponding to the classification loss. In this paper, the weight of each layer is dynamically adjusted to allow the network to have better diagnostic performance in the target domain. This is done by measuring the proportion of each layer's loss to the total loss, and it has been found that the loss value weight of each layer will have a greater effect on the network performance. To balance the three losses of $l_{CLASS}$, $l_{MMD}$ and $l_{CORAL}$ in the training process, this is done. Dynamically adjusting the hyperparameters $x_i, y_i, n$ are shown in equations (8), (9) and (10).

$$x_i = \frac{l_{MMD}^i}{l_{CLASS} + \sum_{i=1}^{t} l_{MMD}^i + \sum_{i=1}^{t} l_{CORAL}^i} \tag{8}$$

$$y_i = \frac{l_{CORAL}^i}{l_{CLASS} + \sum_{i=1}^{t} l_{MMD}^i + \sum_{i=1}^{t} l_{CORAL}^i} \tag{9}$$

$$n = \frac{l_{CLASS}}{l_{CLASS} + \sum_{i=1}^{t} l_{MMD}^{i} + \sum_{i=1}^{t} l_{CORAL}^{i}} \tag{10}$$

By using the three losses mentioned above, the first fully connected layer to be trained can simultaneously reduce the classification loss, the MMD loss, and the CORAL loss. This allows the parameters of the network model to be updated, and the total loss is expressed as the equation (11).

$$l_{LOSS} = n * l_{CLASS} + \sum_{i=1}^{t} x_i l_{MMD}^{i} + \sum_{i=1}^{t} y_i l_{CORAL}^{i} \tag{11}$$

By combining the two losses previously proposed, the classification and CORAL losses can be jointly minimized in the two-layer fully connected layer to modify the parameters of the network model, with the total loss expressed as equation (12).

$$l_{LOSS} = n * l_{CLASS} + \sum_{i=1}^{t} y_i l_{CORAL}^{i} \tag{12}$$

By combining the two losses previously proposed, the classification and MMD losses can be jointly minimized in the two-layer fully connected layer to modify the parameters of the network model, with the total loss expressed as equation (13).

$$l_{LOSS} = n * l_{CLASS} + \sum_{i=1}^{t} x_i l_{MMD}^{i} \tag{13}$$

By giving the CORAL loss a large weight to GAMMA and combining the two losses mentioned above, the classification loss and the MMD loss in the first fully connected layer can be minimized simultaneously, with the MMD loss being given a significant weight, the parameters of the network model being adjusted, and the total loss being expressed as equation (14).

$$l_{LOSS} = l_{CLASS} + GAMMA * l_{CORAL} \tag{14}$$

By giving the MMD loss a large weight to GAMMA and combining the two losses mentioned above, both the classification and MMD losses in the first fully connected layer can be minimized simultaneously, with the MMD loss given a significant weight, the parameters of the network model adjusted, and the total loss expressed as equation (15).

$$l_{LOSS} = l_{CLASS} + GAMMA * l_{MMD} \tag{15}$$

## 5. RESULTS AND DISCUSSIONS

For verifying the effectiveness of multi-layer domain adaptive learning method in predicting disk failure, a comparison is made with a single-layer domain adaptive learning model and a model without domain adaptive learning. As the target field, Figures 2, 3 and 4 display the experimental results, which were initially determined by the disk with the most defective data.

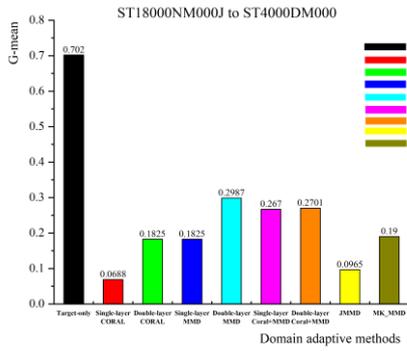
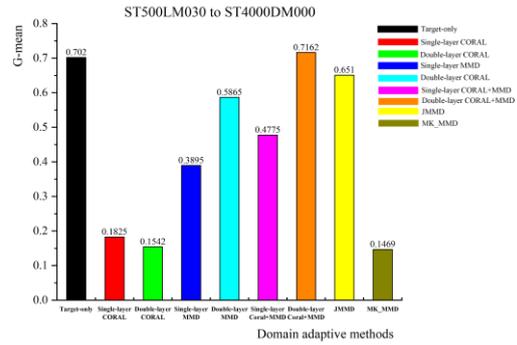

(a) ST14000NM001G to ST4000DM000    (b) ST14000NM001G to ST8000NM0055

Figure 2. ST50OLM030 to ST4000DMO00 and ST18000NM000J to ST4000DM000

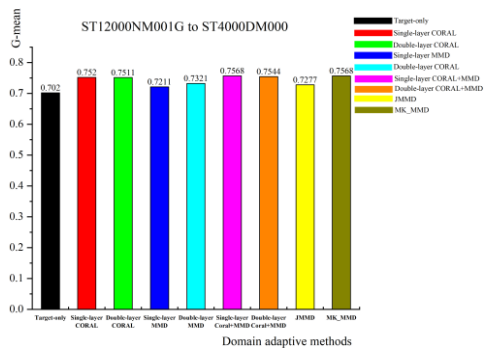
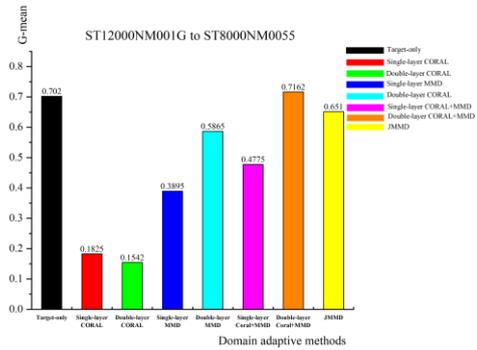

(a) ST12000NMO01G to ST8000NM0055    (b) ST12000NM001G to ST4000DM000

Figure 3. ST12000NMO01G to ST8000NM0055 and ST12000NM001G to ST4000DM000

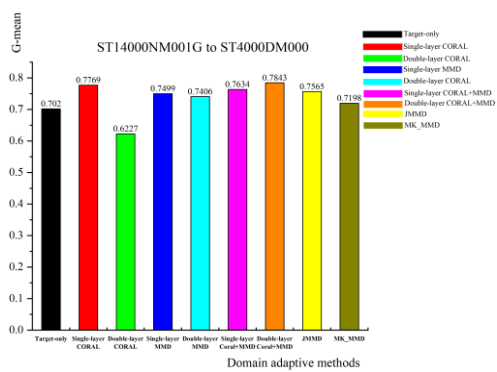
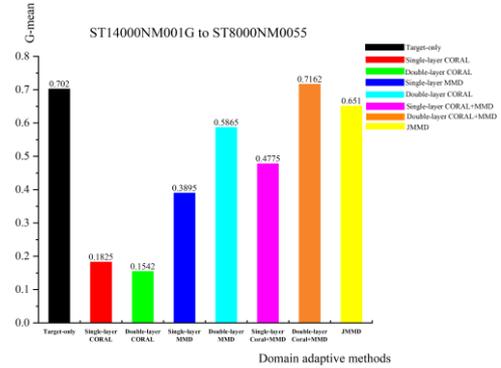

(a) ST14000NM001G to ST400ODM000    (b) ST14000NM001G to ST8000NMO055

Figure 4. ST14000NMO01G to ST4000DMO00 and ST14000NMO01G to ST8000NM0055

Experiments show that a small number of failed disk data disks are not suitable for domain adaptation as the source domain, and negative migration occurs. In contrast to Single-layer and Double-layer CORAL, MMD, Multi-Kernel Maximum Mean Difference (MK_MMD), Joint Maximum Mean Discrepancy (JMMD) and other methods, this paper's proposed multi-layer domain adaptation method still performs better.

Table 3, 4, 5, 6, 7 and 8 display the recognition rates of three disks with a limited number of breakdowns as the target domain.

Table 3: G-mean score for ST10000NM0086 type disk

| Source domain | ST4000DM000 | ST8000DM002 | ST12000NM001G | ST12000NM0007 |
|---|---|---|---|---|
| Target domain | ST10000NM0086 | ST10000NM0086 | ST10000NM0086 | ST10000NM0086 |
| Target-only | 0.7071 | 0.7071 | 0.7071 | 0.7071 |
| Source-only | 0.7388 | 0.5503 | 0.8295 | 0.8575 |
| Single-layer CORAL | 0.8737 | 0.9039 | 0.8478 | 0.8786 |
| Double-layer CORAL | 0.7872 | 0.8478 | 0.8898 | 0.8590 |
| Single-layer MMD | 0.8553 | 0.8704 | 0.7890 | 0.8802 |
| Double-layer MMD | 0.8753 | 0.8891 | 0.8088 | 0.8786 |
| Single-layer Coral+MMD | 0.8959 | 0.8891 | 0.8298 | 0.8802 |
| Double-layer Coral+MMD | 0.8753 | 0.8898 | 0.8332 | 0.8981 |

Table 4 G-mean score for ST10000NM0086 type disk

| Source domain | ST8000NM0055 | ST12000NM0008 | ST14000NM0138 | ST14000NM001G |
|---|---|---|---|---|
| Target domain | ST10000NM0086 | ST10000NM0086 | ST10000NM0086 | ST10000NM0086 |
| Target-only | 0.7071 | 0.7071 | 0.7071 | 0.7071 |
| Source-only | 0.7105 | 0.5503 | 0.7454 | 0.8671 |
| Single-layer CORAL | 0.8773 | 0.8753 | 0.9143 | 0.8357 |
| Double-layer CORAL | 0.8023 | 0.7696 | 0.8347 | 0.8081 |
| Single-layer MMD | 0.8089 | 0.8874 | 0.8755 | 0.8704 |
| Double-layer MMD | 0.8247 | 0.8874 | 0.9126 | 0.8942 |
| Single-layer Coral+MMD | 0.8263 | 0.9143 | 0.9126 | 0.8786 |
| Double-layer Coral+MMD | 0.8474 | 0.8925 | 0.9126 | 0.8959 |

Table 5 G-mean score for ST500LM030 type disk

| Source domain | ST4000DM000 | ST8000DM002 | ST12000NM001G | ST12000NM0007 |
|---|---|---|---|---|
| Target domain | ST500LM030 | ST500LM030 | ST500LM030 | ST500LM030 |
| Target-only | 0.0000 | 0.0000 | 0.0000 | 0.0000 |
| Source-only | 0.3389 | 0.4289 | 0.3524 | 0.3524 |
| Single-layer CORAL | 0.3535 | 0.4316 | 0.3535 | 0.3535 |
| Double-layer CORAL | 0.5994 | 0.4302 | 0.7315 | 0.6519 |
| Single-layer MMD | 0.5434 | 0.4952 | 0.3535 | 0.3535 |
| Double-layer MMD | 0.7071 | 0.4952 | 0.6748 | 0.5768 |
| Single-layer Coral+MMD | 0.7090 | 0.4316 | 0.3535 | 0.7120 |
| Double-layer Coral+MMD | 0.7071 | 0.4905 | 0.6748 | 0.5768 |

Table 6 G-mean score for ST500LM030 type disk

| Source domain | ST8000NM0055 | ST12000NM0008 | ST14000NM001G | ST14000NM0138 |
|---|---|---|---|---|
| Target domain | ST500LM030 | ST500LM030 | ST500LM030 | ST500LM030 |
| Target-only | 0.0000 | 0.0000 | 0.0000 | 0.0000 |
| Source-only | 0.3963 | 0.3524 | 0.3513 | 0.4276 |
| Single-layer CORAL | 0.4841 | 0.3535 | 0.3535 | 0.3535 |
| Double-layer CORAL | 0.4201 | 0.6531 | 0.5430 | 0.5146 |
| Single-layer MMD | 0.4302 | 0.4759 | 0.5236 | 0.5236 |
| Double-layer MMD | 0.3535 | 0.4960 | 0.6789 | 0.6293 |
| Single-layer Coral+MMD | 0.4289 | 0.3535 | 0.5775 | 0.7207 |
| Double-layer Coral+MMD | 0.4330 | 0.4960 | 0.6760 | 0.6293 |

Table 7 G-mean score for ST18000NM000J type disk

| Source domain | ST4000DM000 | ST8000DM002 | ST12000NM001G | ST12000NM0007 |
|---|---|---|---|---|
| Target domain | ST18000NM000J | ST18000NM000J | ST18000NM000J | ST18000NM000J |
| Target-only | 0.0000 | 0.0000 | 0.0000 | 0.0000 |
| Source-only | 0.0000 | 0.0000 | 0.3873 | 0.000 |
| Single-layer CORAL | 0.6892 | 0.6324 | 0.7416 | 0.0000 |
| Double-layer CORAL | 0.7071 | 0.6892 | 0.7071 | 0.5244 |
| Single-layer MMD | 0.7071 | 0.5000 | 0.5916 | 0.7071 |
| Double-layer MMD | 0.7071 | 0.7416 | 0.4183 | 0.5700 |
| Single-layer Coral+MMD | 0.6892 | 0.7071 | 0.6892 | 0.7071 |
| Double-layer Coral+MMD | 0.8944 | 0.7071 | 0.7071 | 0.8062 |

Table 8 G-mean score for ST18000NM000J type disk

| Source domain | ST8000NM0055 | ST12000NM0008 | ST14000NM001G | ST14000NM0138 |
|---|---|---|---|---|
| Target domain | ST18000NM000J | ST18000NM000J | ST18000NM000J | ST18000NM000J |
| Target-only | 0.0000 | 0.0000 | 0.0000 | 0.0000 |
| Source-only | 0.0000 | 0.0000 | 0.0000 | 0.0000 |
| Single-layer CORAL | 0.7071 | 0.7071 | 0.7071 | 0.7416 |
| Double-layer CORAL | 0.7071 | 0.7071 | 0.6324 | 0.7416 |
| Single-layer MMD | 0.7071 | 0.7071 | 0.7071 | 0.6708 |
| Double-layer MMD | 0.7071 | 0.0000 | 0.8366 | 0.5916 |
| Single-layer Coral+MMD | 0.7071 | 0.8062 | 0.7071 | 0.7071 |
| Double-layer Coral+MMD | 0.9219 | 0.9219 | 0.8366 | 0.7071 |

It is clear from the experiments that when the model disk with less fault data is used as the target domain, the domain adaptive learning method is not added, and only training by itself and adding the source domain is performed. It compares the basic prediction effectiveness of the multi-layer domain adaptation proposed in this study with other existing domain adaptation methods across eight different target domains. The results consistently demonstrate that the multi-layer domain adaptation outperforms alternative approaches in terms of predictive performance and it solves the problems of small sample number, insufficient reference number, and low fault prediction performance.

Single-layer domain adaptation only adjusts the features of a single layer, missing useful features extracted by other higher layers. Adopting multi-layer domain adaptation can make full use of more and deeper extracted features. By training the model to extract more deep common features, the two domains can be better aligned, and enhancement of the model's performance in predicting failures can be achieved.

## 5. CONCLUSIONS AND OUTLOOKS

This study delves into the challenge of low failure prediction performance arising from the limited availability of faulty disk data in storage systems. We propose a multi-layer domain adaptive learning method that uses neural networks to extract multi-layer features and fully exploits all extracted features during domain adaptation. By dynamically adjusting the weighting parameters of each layer's loss function, the disparity between the two fields is reduced and the model's predictive performance is improved.

In the next step, we will further explore the use of the relationship between multiple source domains and target domains, as well as the relationship between different source domains, so as to improve the fault prediction performance of the target domain. At the same time, explore the advantages of the cumulative function approaching the step function method and the currently used ReLu activation function method, and try to propose an approximate mathematical model. For example, according to the consideration, analogy with the existing model in the field of debugging and testing theory, using the dynamic difference model to describe etc. and take an explicit mathematical description of the transfer function-generated generic activation function.

Finally, it is fervently hoped that the research in this paper can be utilized in storage systems and bring certain social benefits.

## ACKNOWLEDGEMENTS

The research work in this paper was supported by the Shandong Provincial Natural Science Foundation of China (Grant No. ZR2019LZH003), Science and Technology Plan Project of University of Jinan (No. XKY2078) and Teaching Research Project of University of Jinan (No. J2158). Peng Wu is the author to whom all correspondence should be addressed.

**Authors**

Guangfu Gao received a bachelor's degree in computer science and technology from Yantai University. He is currently pursuing a master's degree in computer technology from Jinan University. His research interests include industrial data processing and failure prediction.

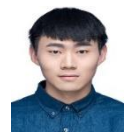

Peng Wu received Master from University of Jinan and a doctoral degree from Beijing Normal University.His research interests include pattern recognition, bioinfor-matics, intelligent computing theory and application research,etc

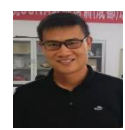

Hussain Dawood received a Ph.D. and a Master's degree from Beijing Normal University. His research interests include object detection, sentiment analysis and pattern recognition, etc.

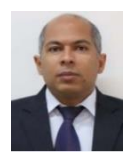